\documentclass[conference]{IEEEtran}
\usepackage{times}

\usepackage[numbers]{natbib}
\usepackage{multicol}
\usepackage[bookmarks=true]{hyperref}
\usepackage{graphicx}
\usepackage{float}
\usepackage{xcolor}

\newcommand{\aggiungi}[1]{\textcolor{black}{#1}}

\begin{document}

\title{Enhancing Graph Representation of the Environment through Local and Cloud Computation}


\author{\authorblockN{Francesco Argenziano, Vincenzo Suriani, and Daniele Nardi}
    \authorblockA{Dept. of Computer, Control, and Management Engineering, Sapienza University of Rome,
    Italy. \\ \{lastname\}@diag.uniroma1.it}}


%

\maketitle

\begin{abstract}
Enriching the robot representation of the operational environment is a challenging task that aims at bridging the gap between low-level sensor readings and high-level semantic understanding. Having a rich representation often requires computationally demanding architectures and pure point cloud based detection systems that struggle when dealing with everyday objects that have to be handled by the robot. To overcome these issues, we propose a graph-based representation that addresses this gap by providing a semantic representation of robot environments from multiple sources. In fact, to acquire information from the environment, the framework combines classical computer vision tools with modern computer vision cloud services, ensuring computational feasibility on onboard hardware. By incorporating an ontology hierarchy with over 800 object classes, the framework achieves cross-domain adaptability, eliminating the need for environment-specific tools. The proposed approach allows us to handle also small objects and integrate them into the semantic representation of the environment. The approach is implemented in the Robot Operating System (ROS) using the RViz visualizer for environment representation. This work is a first step towards the development of a general-purpose framework, to facilitate intuitive interaction and navigation across different domains. 
\end{abstract}

\IEEEpeerreviewmaketitle

\section{Introduction}

\begin{figure}[t]
    \centering
    \includegraphics[width=1.0\columnwidth]{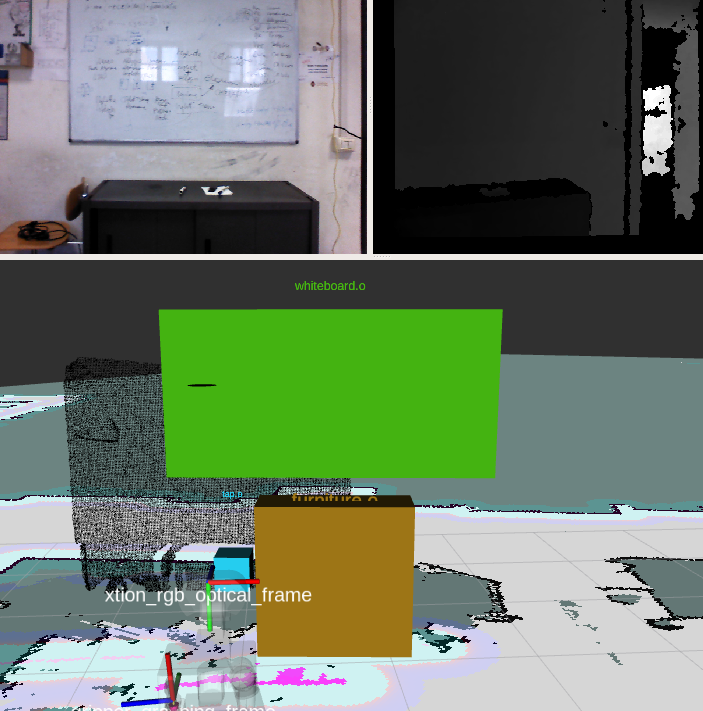}
    \caption{The obtained 3D scene representation in RViz with the exploration of a small portion of the environment. The objects are delimited by the bounding boxes (obtained from the cloud computation) fused with the point cloud information. After this, the corresponding graph representation is generated.}
    \label{fig:intro}
\end{figure}

\begin{figure*}[t]
    \centering
    \includegraphics[width=2.0\columnwidth]{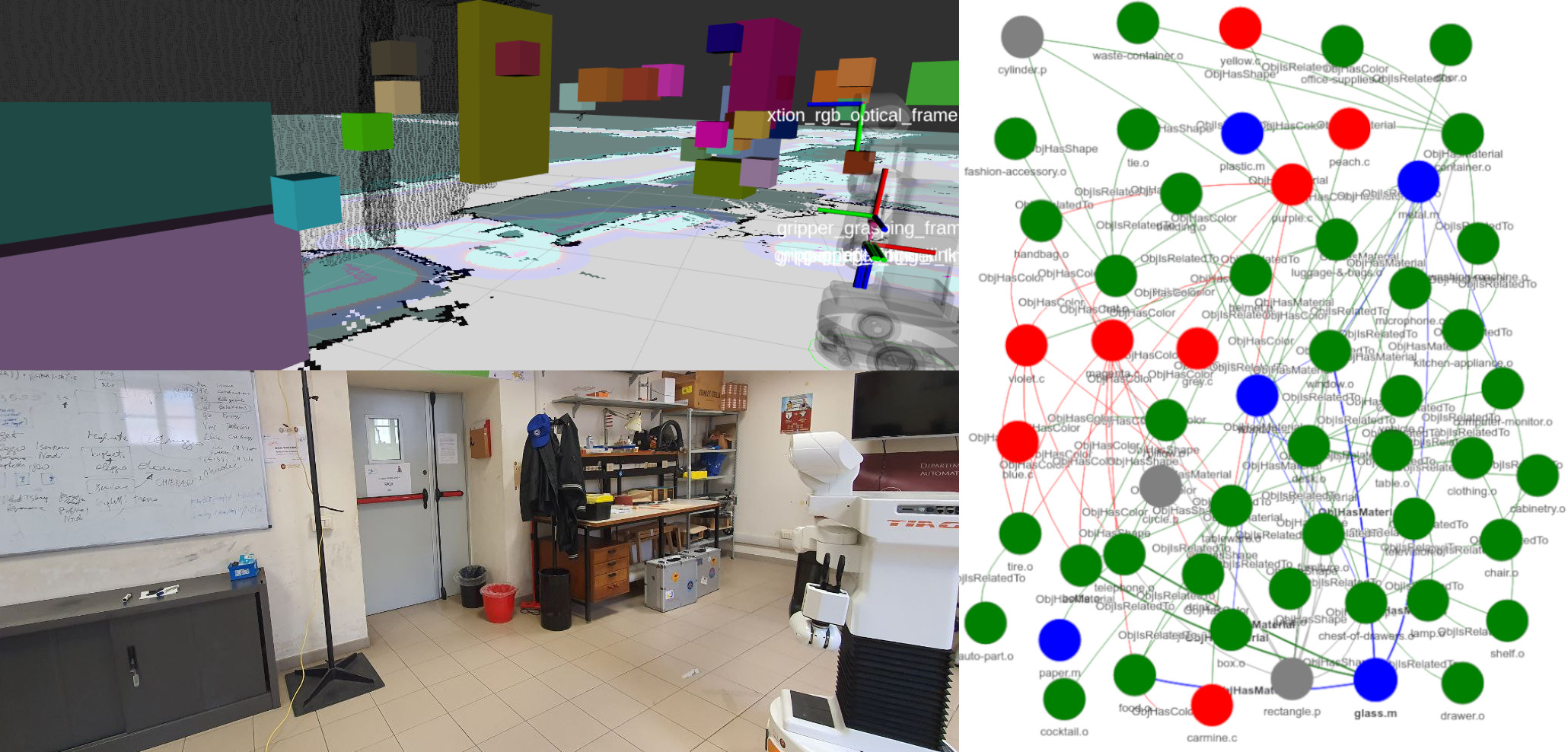}
    \caption{On the left: population in the RViz visualizer of the objects detected thanks to Google Cloud's API. On the right, the graph extracted from the underlying relations between objects, and between objects and their properties. Different colors mean different semantic groups of the nodes: materials, shapes, objects and colors.}
    \label{fig:intro_banner}
\end{figure*}

In recent years, the field of robotics has witnessed significant advancements in perception capabilities, thanks to the proliferation of sensors and computer vision techniques. However, bridging the gap between low-level sensor readings and high-level semantic understanding remains a challenge. To this end, we propose a framework, in its early stages, 
that tackles this gap using a graph representation to connect sensor data with a semantic representation of the environment.

To be able to have good precision in object detection and achieve computationally acceptable performance on resource-constrained robotic hardware, our approach combines classical computer vision tools with modern computer vision cloud services. By leveraging the power of cloud computing, we can offload intensive processing tasks and ensure real-time responsiveness even on limited onboard hardware. Since robots often need to deal with small objects in the environment, we adopted the cloud-vision system to have high precision on single objects in the environment.

One of the major advantages of our framework lies in its ability to be cross-domain and adaptable to different environments without requiring the development of specific customizations for each scenario. This is achieved through the incorporation of an ontology hierarchy, encompassing more than 800 object classes. We integrated the remote hierarchy with the local ontology to obtain a unified graph representation that can be used in the robot's tasks. By using such a comprehensive ontology, our framework can handle diverse environments and objects, facilitating seamless navigation and interaction across various contexts.

To semantically represent the robot's environment, we employ a set of entity classes to define objects and their attributes, while a graph representation is utilized to establish connections between the environment's entities. \aggiungi{Such representation can be then exploited also to perform many Human-Robot Interaction tasks, for example in unexplored environments. Another scenario of application is in exploring and storing the information of a scene at different time frames to capture the temporal evolution of an environment (ideally, humans could ask the robotic agent information related to a specific time instance, like when the room was tidied rather than the actual version of the environment) and this can also be studied within the Continual Learning framework.}

We implement the proposed architecture on the Robot Operating System (ROS) and use the RViz visualizer, allowing for intuitive visualization and interaction with the environment. An example of this visualization can be seen in Fig. \ref{fig:intro}. The platform used is the TIAGo robot, manufactured by PAL Robotics\footnote{https://pal-robotics.com/}.

The rest of this paper is organized as follows: Section 2 provides a brief overview of related work in 
semantic representation in robotics. Section 3 elaborates on our proposed framework, highlighting the graph-based representation and ontology hierarchy
, focusing on the integration with ROS and the chosen platform. Section 4 presents concluding remarks.

\section{Related Work}

Semantic representation in robotics is a vital area of research, enabling robots to comprehend and interact with their environment effectively. Various approaches have been proposed to bridge the gap between sensor data and semantic understanding. Object recognition algorithms and scene understanding techniques \cite{gupta2014} are commonly employed to extract high-level semantic information from sensor data, facilitating intelligent decision-making by robots.  

In last years, graph-based approaches have gained popularity in robotics due to their ability to capture complex relationships and dependencies within the environment. By representing the environment as a graph, these frameworks provide a structured representation that facilitates semantic understanding and reasoning. Graph-based frameworks have been successfully applied to object detection \cite{zhang2018} and scene parsing \cite{li2019}, enabling robots to perceive and interpret their surroundings effectively. 3D scene graphs have been presented and used in \cite{3dscenegraph}, \cite{3dscenegraph2} to represent 3D environments. In those, nodes represent spatial concepts at multiple levels of abstraction and edges represent relations between concepts. One of the limitations in building such a representation automatically has been represented by the computational costs.  To overcome such issues, recently, in \cite{hydra_carlone} the authors introduced Hydra, capable to build incrementally a 3D scene graph from sensor data in real time thanks to the combination of novel online algorithms and a highly parallelized perception architecture. Another approach, capable of incrementally building the scene graph but also aggregating PointNet\cite{pointnet} features from primitive scene components using graph neural network has been proposed in \cite{tombari}. In this, an attention mechanism has also been proposed to deal with missing graph data in incremental reconstruction scenarios. 
When dealing with local approaches, the set of objects that are detectable is quite limited due to the limited availability in hosting large neural network models. To this end, in order to guarantee cross-domain adaptability through a large set of detectable objects and computational sustainability on robot CPUs, cloud services have been adopted as an addition to the local perception pipeline.  
To address this challenge, cloud services for computer vision, such as Google Cloud Vision\footnote{cloud.google.com/vision}, have emerged as viable solutions in robotic applications \cite{7556193}. By leveraging cloud services, robots can offload processing tasks, enabling real-time perception even on resource-constrained hardware \cite{mello2022introduction}.
By relying on this platform we fused the remote hierarchy with the local ontology 
to map the environment obtaining a unified graph representation that can be used for robot navigation and object localization tasks.
\aggiungi{Moreover, with respect to the state of the art, we are working on a different level of detail. In this way, we are able to include in the scene representation a wider variety of objects, thus expanding the set of future possible human-robot interaction scenarios.}


\section{Methodology}

\begin{figure}[t]
    \centering
    \includegraphics[width=1.0\columnwidth]{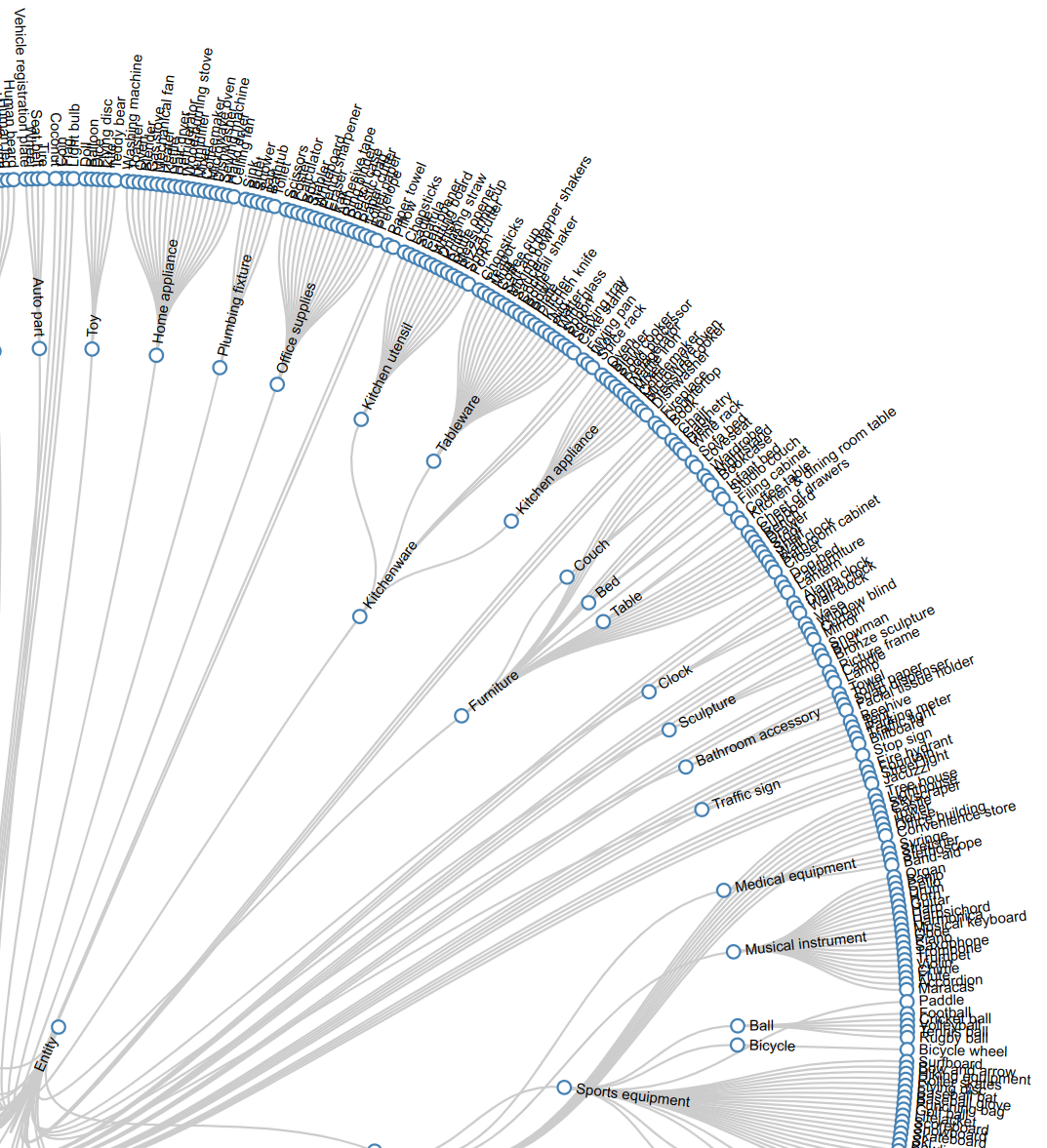}
    \caption{Part of the hierarchy class from Google Cloud Vision. From the entity, the father of all the entities, more than 800 classes are provided. The full set of classes is available at \url{ https://storage.googleapis.com/openimages/2018_04/bbox_labels_600_hierarchy_visualizer/circle.html}}
    \label{fig:ontology}
\end{figure}

\subsection{Entity Representation and Hierarchy}
\begin{figure*}[t]
    \centering
    \includegraphics[width=1.9\columnwidth]{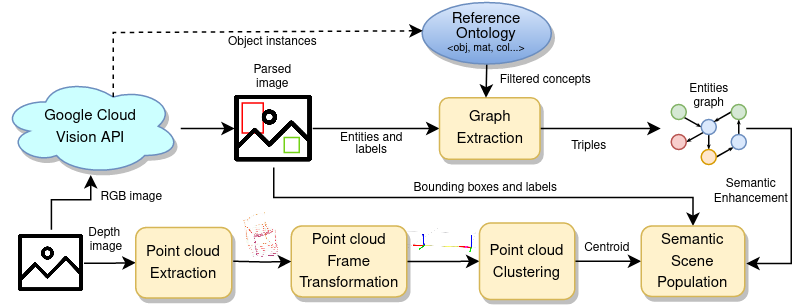}
    \caption{The pipeline presented in this work. The local and remote perception branches lead to a unified graph representation of the environment.}
    \label{fig:architecture}
\end{figure*}

Dealing with hierarchies of concepts with too many entries can be very difficult to handle, especially when cross-domain applications are involved. Hierarchical ontologies that include very deep levels of details can be composed of thousands of classes, thus there is a bigger risk of miss-classifying the objects in the real world that the robotic agent can come across. Choosing an adequate level of detail is therefore a key aspect for robots that are set to explore different domains. It is desired that these agents should be able to recognize objects and properties of such objects (for example, shapes and materials) that an average human participant in the interaction should recognize. To this end, the proposed approach exploits Google Cloud Vision APIs and their underlying taxonomy. The main idea behind this choice is that for a robotic agent that should eventually enter in contact with some humans, is not requested to deal with very detailed concepts that the human agent could not be aware of, and therefore an ontology hierarchy based on everyday-life objects and entities (even if cross-domain) is enough. This taxonomy can be observed in Fig. \ref{fig:ontology}. Such hierarchy is composed as follows:
\begin{itemize}
    \item the root is labelled as \textit{Entity}, from which every other concept is derived;
    \item the first level of descendants consists on some basic general concepts that group several categories like \textit{Animal}, \textit{Vehicle}, \textit{Building} and so on;
    \item from this point, the hierarchy is refined in a non-homogeneous fashion, with some classes refined more times than others before reaching the leaf concepts of the tree; 
    \item at the end of the hierarchy there are the most basic concepts, like \textit{Hammer}, \textit{Dishwasher} and \textit{Bee}.
\end{itemize}
For the purposes of this research, classes and concepts concerning animals and people were removed. Despite this, the resulting taxonomy is still composed of more than 800 classes, which we considered a suitable amount for this application. 

In addition to the main class, Google Vision APIs are able to provide of additional boundary information of the detected object. The challenge with this information is that comes all together and in an unstructured manner. It is important that concepts with different semantics (e.g. the object class and the color, or the material) are treated differently. To this end, we extended the labels provided by Google by reorganizing them in an ad-hoc augmented ontology: in this ontology, every concept is represented in the format \textit{'name.e'}, in which the 'e' letter changes depending on what is the semantic on the information we are representing. We have distinguished 4 main groups of concepts: \textit{'.o'} stands for \textit{objects} (i.e. 'hat.o'); \textit{'.m'} stands for material ('plastic.m'); \textit{'.s'} stands for shape ('cube.s'); \textit{'.c'} stands for color ('red.c'). With this organization, we are also able to express relations between concepts: a detected blue chair in the environment is translated as the triple "chair.o ObjHasColor blue.c"; and since the same object can hold multiple relations ("chair.o ObjHasColor blue.c" but also "chair.o ObjHasMaterial plastic.m"), this means that even few images of the environment can produce numerous triples that are then stored and represented within a knowledge graph. \aggiungi{All the possible features that are assigned to the objects of the world can come from different sources. The ones that are hard to perceive locally by the robot (like materials of objects) are extracted entirely in cloud, while others are obtained by generalizing and abstracting from other objects using an approach derived by \cite{bartoli2022knowledge}.}
An example of the output of the system can be seen in Fig. \ref{fig:intro_banner}, while the architecture can be observed in Fig. \ref{fig:architecture}.
\subsection{ROS Architecture}
In the lower portion of the Fig. \ref{fig:architecture}, it is possible to observe the pipeline of operations that involves the depth images that come from the TIAGo robot's RGB-D camera. Initially, from each image, the more relevant point clouds are extracted and referenced. These point clouds, however, are expressed in the robot reference frame and therefore are projected in the map reference frame. Then, from these point clouds the respective clusters are computed using the Open3D library, and for each cluster, a centroid is computed, in order to have a single point of reference for each of them. Finally, a correspondence between the bounding boxes detected by Google's APIs and the computed centroid is made in order to keep only the meaningful objects for that iteration. Such objects are then projected in RViz with apposite markers and, in the end, thanks to the information that comes from the knowledge graph, the markers are linked in order to form a topological graph of the objects in the environment.

\section{Conclusion} 
\label{sec:conclusion}

In this paper, we have presented a first step in building a software system that fuses classical computer vision tools and cloud computer vision services to obtain a semantic representation of the environment. We have ensured computational feasibility for onboard hardware and high performance in object detection accuracy and capability. Using a graph-based approach that comprehends the full set of objects of the Google Cloud Vision detector, we integrate an ontology hierarchy, consisting of 800+ object classes. This has enabled cross-domain adaptability and eliminated the need for developing specific tools for different environments.


This is a first step in the implementation of a blended system, since, for example, there is plenty of room to improve the mapping between bounding boxes classified from the cloud service and the mapping on the point cloud depth. In fact, the majority of the issues come when dealing with partial occlusions in the detected objects. 
To this end, we aim to improve the integration between the local representation and the cloud-computed one, taking into account not only clusters of point clouds but preserving the shapes obtained from them. We also plan to explore the integration of machine learning techniques to improve the accuracy and robustness of semantic understanding in dynamic environments.

Overall, the proposed approach opens up avenues for enhanced perception and interaction capabilities of robots in various domains. By leveraging the power of semantic representation, we envision a future where robots can understand and navigate complex environments, enabling seamless human-robot interaction and collaboration in diverse settings.

\section*{Acknowledgments}
We acknowledge partial financial support from PNRR MUR project PE0000013-FAIR.
This work has been carried out while Francesco Argenziano was enrolled in the Italian National Doctorate on Artificial Intelligence run by Sapienza University of Rome.

\bibliographystyle{plainnat}
\bibliography{references}

\end{document}